\documentclass{ifacconf}

\usepackage{graphicx}     
\usepackage{natbib}       
\usepackage{amsmath}
\usepackage{amssymb}  
\usepackage[ruled, vlined, linesnumbered, rightnl, algo2e, norelsize]{algorithm2e} 
\usepackage{subcaption}
\usepackage{stfloats}
\usepackage{textpos}
\begin{document}
\begin{frontmatter}

\begin{textblock*}{20cm}(-4.8cm, -0.3cm)
	\textbf{\footnotesize $\dagger$ Published in IFAC-PapersOnLine, Volume 54, Issue 21, Pages 157--162, 2021, Elsevier.}
\end{textblock*}
\begin{textblock*}{20cm}(-8.7cm, 0.1cm)
	\textbf{\footnotesize DOI: 10.1016/j.ifacol.2021.12.027}
\end{textblock*}
\begin{textblock*}{10cm}(2.5cm, 25.5cm)
	\textbf{\footnotesize Copyright \copyright \space 2021 IFAC}
\end{textblock*}

\title{Optimised Informed RRTs for Mobile Robot Path Planning} 

\author{Bongani B. Maseko,} 
\author{Corn\'{e} E. van Daalen,} 
\author{Johann Treurnicht}

\address{Electronic Systems Laboratory, Department of Electrical \& Electronic Engineering, Stellenbosch University, South Africa}

\begin{abstract}               
Path planners based on basic rapidly-exploring random trees (RRTs) are quick and efficient, and thus favourable for real-time robot path planning, but are almost-surely suboptimal. In contrast, the optimal RRT (RRT*) converges to the optimal solution, but may be expensive in practice. Recent work has focused on accelerating the RRT*'s convergence rate. The most successful strategies are informed sampling, path optimisation, and a combination thereof. However, informed sampling and its combination with path optimisation have not been applied to the basic RRT. Moreover, while a number of path optimisers can be used to accelerate the convergence rate, a comparison of their effectiveness is lacking. This paper investigates the use of informed sampling and path optimisation to accelerate planners based on both the basic RRT and the RRT*, resulting in a family of algorithms known as \emph{optimised informed RRTs}. We apply different path optimisers and compare their effectiveness. The goal is to ascertain if applying informed sampling and path optimisation can help the quick, though almost-surely suboptimal, path planners based on the basic RRT attain comparable or better performance than RRT*-based planners. Analyses show that RRT-based optimised informed RRTs do attain better performance than their RRT*-based counterparts, both when planning time is limited and when there is more planning time.
\end{abstract}


\end{frontmatter}

\section{Introduction}

Key requirements for an autonomous transportation system include safety, reliability and responsiveness. The path planner, which is a critical component of such a system, therefore has to plan near-optimal paths that result in reaching destinations safely. Moreover, paths must be planned quickly, otherwise unacceptable delays (in real-time path planning) are incurred. The planned path then functions as the reference input to the vehicle controller, which has the task of accurately following the path.

The most widely used path-planning approaches are based on the configuration space~\citep{Lozano_1983}; they include reactive, combinatorial, grid-based, potential field, decision-theoretic, and sampling-based path planning~\citep{Lumelsky_1987, Chazelle_1987, Yap_2002, Hwang_1992, Du_2010, Kavraki_PRM, LaValle_RRT}. We use sampling-based path planners, which are the most established; they efficiently generate good quality paths and can seamlessly incorporate vehicle motion constraints. We focus on single-query applications, where a robot plans to navigate an environment only once. The basic rapidly-exploring random tree (RRT)~\citep{LaValle_RRT} is a quick and efficient single-query sampling-based path planner that has, unfortunately, been proven to be almost-surely suboptimal. The optimal RRT (RRT*)~\citep{Karaman_RRTStar}, an asymptotically-optimal extension of the RRT, finds a solution that tends to the optimal solution as the number of iterations tends to infinity; so, it may take very long to find a near-optimal path. After the introduction of the RRT*, some work has focused on accelerating its convergence rate, with the aim of finding a near-optimal path in finite time. 

Although some strategies, notably informed sampling \citep{Gammell_2014} and its combination with path optimisation~\citep{Kim_2015}, have been shown to be effective in accelerating the RRT*'s convergence, this has not been compared to similar acceleration of the basic RRT. Basic RRT-based planners are usually much faster than RRT*-based planners, which would allow more time to improve the solution. This then raises the question of whether applying informed sampling and path optimisation can help the quick, though almost-surely suboptimal, path planners based on the basic RRT attain comparable or better performance than their RRT*-based counterparts. In literature, the effectiveness of different path optimisers in accelerating convergence has not been studied. To this end, we apply a combination of informed sampling and path optimisation to both planners based on the basic RRT and the RRT*, and compare their convergence rates. We use different path optimisers, including path pruning, random shortcut, the wrapping process and gradient-based path optimisation~\citep{Geraerts_2007, Kim_2015, Campana_2016}.

In this work, the algorithms are presented in their most basic form, wherein planned paths are piecewise linear. Such paths are commonly used for robots with no significant differential constraints or those with an onboard controller that is capable of tracking such paths without sustaining collisions. Adaptation for robots with significant differential constraints is part of future work. 

The paper is organised as follows: The benchmark basic RRT and RRT* that use informed sampling are presented in Section~\ref{sec:Benchmark_Path_Planners}, followed by the path optimisers and their application to the benchmark planners in Section~\ref{sec:Path_Optimisers}.  Comparative results are presented in Section~\ref{sec:Accelerated_Path_Planners}, followed by conclusions in Section~\ref{sec:Conclusion}.

\section{Benchmark Path Planners}
\label{sec:Benchmark_Path_Planners}

We start this section by reviewing the basic RRT and explain 
how the RRT* extends it. Thereafter, focus shifts to their informed versions -- the benchmark path planners.

\subsection{Basic RRT and RRT*}

The basic RRT grows a planning tree, $\mathcal{T}$, rooted at the initial configuration, $\mathbf{q}_I$, that explores the configuration space, $\mathcal{C}$. Its pseudocode is given in Alg.~\ref{alg:RRT}. The tree is grown by iteratively sampling a random configuration, $\mathbf{q}_\mathrm{samp}
$, and attempting to extend the tree towards it (lines~7--9). Tree nodes are reachable configurations and its edges are feasible paths by which the configurations are reached. To grow the tree, an existing node, $\mathbf{q}_\mathrm{nearest}$, that is closest to $\mathbf{q}_\mathrm{samp}
$ is selected (line~2). The \texttt{newConfig} function (line~3) then generates a configuration,  $\mathbf{q}_\mathrm{new}$, that gets as close as possible to $\mathbf{q}_\mathrm{samp}
$ through a feasible and collision-free edge from $\mathbf{q}_\mathrm{nearest}$. The edge is computed using a local path-planning method (LPM) and tested for collisions using a collision detector before $\mathbf{q}_\mathrm{new}$ is added. This is performed by the \texttt{insertNode} function (line~4).
Once a node within a specified distance from the goal configuration, $\mathbf{q}_G$, is added, a solution path is considered found; alternatively the goal may be occasionally sampled, and an attempt to add it made. Since the basic RRT always extends the tree from the nearest existing node, regardless of its cost (see Fig.~\ref{fig:benchmarkPathPlannersIllustration}(a)), it does not attempt to improve path cost.

\begin{algorithm2e}[h]
\caption{\texttt{RRT}($\mathbf{q}_I,\mathbf{q}_G$)}
\label{alg:RRT}
\SetKwFunction{extendRRT}{\texttt{extendRRT}}
\SetKwFunction{main}{\texttt{main}}
\SetKwProg{myFunc}{Function}{}{}
\SetArgSty{}
\SetFuncSty{}
\small

\myFunc{\extendRRT{$\mathcal{T},\; \mathbf{q}_\mathrm{samp}
$}}{
	$\mathbf{q}_\mathrm{nearest} \gets$ \texttt{nearestNeighbour}($\mathcal{T},\; \mathbf{q}_\mathrm{samp}
$)\;
	\If{\texttt{newConfig}$(\mathbf{q}_\mathrm{nearest}, \mathbf{q}_\mathrm{samp}
,\; \mathbf{q}_\mathrm{new})$}{
		$\mathcal{T}$\texttt{.insertNode}($\mathbf{q}_\mathrm{nearest},\; \mathbf{q}_\mathrm{new}$)\;		
	}
}

\myFunc{\main{}}{
$\mathcal{T}$\texttt{.initialise}($\mathbf{q}_I$)\;
	\For{$i = 1 : \texttt{maxIterations}$}{
			$\mathbf{q}_\mathrm{samp}
 \gets$ 	\texttt{randomConfig}()\; 
    		\texttt{extendRRT}($\mathcal{T},\; \mathbf{q}_\mathrm{samp}
$)\; 
	} 
	\textbf{return} $\mathcal{T}$\; 
}
\end{algorithm2e}

The RRT* is an extension of the RRT that refines the tree with every added node, so as to improve paths to the new node and its neighbours. Its pseudocode is given in Alg.~\ref{alg:RRTStar}. Firstly, instead of simply connecting a new node via the nearest existing node, it considers multiple nodes within a specified radius and chooses the cheapest as the new node's parent (lines~5--7). Secondly, it considers the new node as a replacement parent for neighbouring nodes if reaching such a node via the new node results in a cheaper path than via the current parent. This is known as rewiring (line~8). The RRT* is asymptotically optimal, which means that the solution tends to the optimal solution as the number of iterations tends to infinity. However, in practice it may take very long to find a near-optimal path. This is, in part, due to sampling globally throughout the planning period~\citep{Gammell_2014}. As such, through tree refinement, it wastes time improving paths to every configuration in the planning domain.

\begin{algorithm2e}[h]
\caption{\texttt{RRT}*($\mathbf{q}_I, \mathbf{q}_G$)}
\label{alg:RRTStar}
\SetArgSty{}
\SetFuncSty{}
\SetNoFillComment
\SetKwFunction{extendAndRewireRRTStar}{\texttt{extendAndRewireRRT}*}
\SetKwFunction{main}{main}
\SetKwProg{myFunc}{Function}{}{}
\small

\myFunc{\extendAndRewireRRTStar{$\mathcal{T},\; \mathbf{q}_\mathrm{samp}
$}}{
	$\mathbf{q}_\mathrm{nearest} \gets$ \texttt{nearestNeighbour}($\mathcal{T},\; \mathbf{q}_\mathrm{samp}
$)\;
	$\mathbf{q}_\mathrm{new} \gets$ \texttt{steer}($\mathbf{q}_\mathrm{nearest}, \mathbf{q}_\mathrm{samp}$)\; 
	\If{\texttt{collisionFree}($\mathbf{q}_\mathrm{nearest}, \mathbf{q}_\mathrm{new}$)}{
		$\mathbf{Q}_\mathrm{near} \gets$ \texttt{near}$(\mathcal{T}, \mathbf{q}_\mathrm{new}, r_{\mathrm{RRT}^*})$\;
		$\mathbf{q}_\mathrm{min} \gets$ \texttt{chooseParent}$(\mathbf{Q}_\mathrm{near}, \mathbf{q}_\mathrm{nearest}, \mathbf{q}_\mathrm{new})$\;
		$\mathcal{T}.$\texttt{insertNode}$(\mathbf{q}_\mathrm{min}, \mathbf{q}_\mathrm{new})$\;
		$\mathcal{T}$.\texttt{rewire}$(\mathbf{Q}_\mathrm{near}, \mathbf{q}_\mathrm{min}, \mathbf{q}_\mathrm{new})$\;
	}
}

\myFunc{\main{}}{
	$\mathcal{T}$\texttt{.initialise}($\mathbf{q}_I$)\;
	\For{$i = 1 : \texttt{maxIterations}$}{ 
		$\mathbf{q}_\mathrm{samp} \gets$ 	\texttt{randomConfig}()\; 
		\texttt{extendAndRewireRRT}*$(\mathcal{T},\; \mathbf{q}_\mathrm{samp}$)\;
	}
	\textbf{return} $\mathcal{T}$\; 
}
\end{algorithm2e} 

Since the basic RRT does not refine the tree, it is generally much faster than the RRT*, but it does not attempt to improve path cost. The RRT* improves path cost, but does so inefficiently. Benchmark path planners, which improve these deficiencies, are presented next. We begin with the RRT*-based one, since it exists (as is) in literature.

\subsection{Informed RRT*}
\label{sec:Informed_RRTStar}

The informed RRT*~\citep{Gammell_2014} adapts the RRT* for informed sampling. Its pseudocode is given in Alg.~\ref{alg:informedRRTStar}. It works exactly like the RRT* until the first solution is found. It draws samples from an informed subset of $\mathcal{C}$ -- the subset of configurations that could possibly improve the current solution -- that is determined by the cost of the current best solution, $c_\mathrm{best}$ (line~3); with $c_\mathrm{best}$  infinite, samples are drawn globally. This ensures that tree extension and rewiring (line~4) are only performed for configurations that can lead to a better path.

\begin{algorithm2e}[h]
\caption{\texttt{informedRRT}*($\mathbf{q}_I, \mathbf{q}_G$)}
\SetArgSty{}
\SetFuncSty{}
\label{alg:informedRRTStar}
\small
$\mathcal{T}$\texttt{.initialise}($\mathbf{q}_I$); $\;c_\mathrm{best} \gets \infty$\;
\For{$i \gets 1\; :\; \texttt{maxIterations}$} {
	$\mathbf{q}_{\mathrm{samp}} \gets$ \texttt{informedSample}$(\mathbf{q}_I,\mathbf{q}_G,c_{\mathrm{best}})$\;
	\texttt{extendAndRewireRRT}*$(\mathcal{T},\mathbf{q}_{\mathrm{samp}})$\;
	\If{new solution found}{
		$c_{\mathrm{best}} \gets \mathrm{min}(c_{\mathrm{best}},\;\texttt{cost}(\mathrm{new\; solution}))$\; 
	}
}
\textbf{return} $\mathcal{T}$\;
\end{algorithm2e}  

\subsection{Informed RRT}
\label{sec:Informed_RRT}

An informed version of the basic RRT does not exist in literature; the closest existing algorithm is the anytime RRT~\citep{Fegurson_2006}. We adapt this algorithm for informed sampling to form the \emph{informed RRT}, whose pseudocode is given in Alg~\ref{alg:informedRRT}. The algorithm considers a specified number of neighbouring nodes, $k > 0$, for extension and selects the node that results in the lowest cost from root as the new node's parent (lines~2--7). This results in two modes of operation. In the first mode, which results from setting $k=1$, tree extension is exactly the same as in the basic RRT. Setting $k>1$ results in the second mode, which selects the cheapest among the $k$ neighbouring nodes for tree extension. These modes correspond to the two extremes of the anytime RRT~\citep{Fegurson_2006}. In the first extreme, the anytime RRT is quick, but very suboptimal, and in the second, it is slower, but finds better paths. The informed RRT differs from the anytime RRT in that instead of using rejection sampling to grow the tree, it directly samples the informed subset, just like the informed RRT* (lines~8--13). It differs from the informed RRT* in that while the informed RRT* maintains the same tree throughout the planning period, it grows each tree from scratch (lines~14--18), just like the anytime RRT. We call the first mode the \emph{informed basic RRT} (IB-RRT) and the second, the \emph{informed k-nearest RRT} (IKN-RRT); they serve as benchmark path planners, with the informed RRT* (I-RRT*) being the third. Due to the absence of tree rewiring in the informed RRT, although both its modes are generally faster to find new feasible paths than the informed RRT*, they fail to find a near-

\begin{algorithm2e}[t]
\caption{\texttt{informedRRT}($\mathbf{q}_I, \mathbf{q}_G$)}
\SetArgSty{}
\SetFuncSty{}
\label{alg:informedRRT}
\SetKwFunction{main}{main}
\SetKwFunction{extendInformedRRT}{extendInformedRRT}
\SetKwFunction{growInformedRRT}{growInformedRRT}
\SetKwProg{myFunc}{Function}{}{}
\small
$\mathcal{T} \gets \emptyset$; $\;c_\mathrm{best} \gets \infty$\;

\myFunc{\extendInformedRRT{$\mathcal{T},\; \mathbf{q}_\mathrm{samp}$}}{
	$\mathbf{Q}_\mathrm{near} \gets$ \texttt{kNearestNeighbours}$(\mathcal{T}, \mathbf{q}_\mathrm{samp}, k)$\;
	\While{$\mathbf{Q}_\mathrm{near} \neq \emptyset$}{
		$\mathbf{q}_\mathrm{parent} \gets \mathrm{pop\; out\; cheapest\;  neighbour\; from\; \mathbf{Q}_\mathrm{near}}$\;
		\If{\texttt{newConfig}$(\mathbf{q}_\mathrm{parent},\; \mathbf{q}_\mathrm{samp},\; \mathbf{q}_\mathrm{new})$}{
			$\mathcal{T}$\texttt{.insertNode}($\mathbf{q}_\mathrm{parent},\; \mathbf{q}_\mathrm{new}$)\;	
		}
	}
}

\myFunc{\growInformedRRT{$\mathcal{T},\; c_\mathrm{best}$}}{
	\While{$\texttt{elapsedTime} < \texttt{maxTimePerTree}$}{
		$\mathbf{q}_{\mathrm{samp}} \gets$ \texttt{informedSample}$(\mathbf{q}_I,\mathbf{q}_G,c_{\mathrm{best}})$\;
		\texttt{extendInformedRRT}$(\mathcal{T},\mathbf{q}_{\mathrm{samp}})$\;
		\If{new solution found}{
			$c_{\mathrm{best}} \gets \mathrm{min}(c_{\mathrm{best}},\; \texttt{cost}(\mathrm{new\; solution}))$
		}
	}
}

\myFunc{\main{}}{
	\While{!\texttt{planningTimeElapsed}()}{
		$\mathcal{T}$\texttt{.reInitialise}($\mathbf{q}_I$)\;
		\texttt{growInformedRRT}($\mathcal{T},\; c_\mathrm{best}$)\;
	}
	\textbf{return} $\mathcal{T}$\;
}
\end{algorithm2e}

\begin{figure*}[b]
\centering
	\subfloat[][Informed basic RRT (IB-RRT), $c_\mathrm{best} = 154.8$~m (15$^{th}$ tree).]{%
		\includegraphics[scale = 0.95]{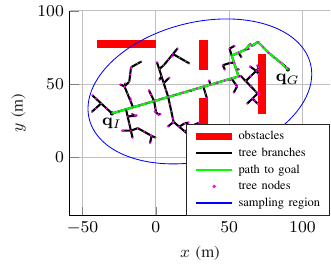}
   	}%
   	\hspace{12.5mm}%
   	\subfloat[][Informed $k$-nearest RRT (IKN-RRT), $c_\mathrm{best} = 142.8$ m (8$^{th}$ tree).]{%
		\includegraphics[scale = 0.95]{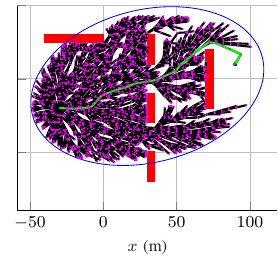}
   	}%
   	\hspace{12.5mm}%
   	\subfloat[][Informed RRT* (I-RRT*),\\ $c_\mathrm{best} = 132.0$ m (4$^{th}$ search).]{%
		\includegraphics[scale = 0.95]{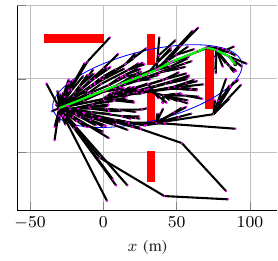}
   	}
\captionsetup{justification = centering}
\caption{Resulting trees when the benchmark path planners are given up to 375 seconds to find a near-optimal path. Since the informed RRT grows each tree from scratch, the tree with the best solution is shown.}
\label{fig:benchmarkPathPlannersIllustration}
\end{figure*}

optimal path, even with more planning time (see Fig.~\ref{fig:benchmarkPathPlannersIllustration}).

As noted by~\citet{Kim_2015}, the convergence rate can be improved by optimising each new path before using its cost for informed sampling. This is the premise for \emph{optimised informed RRTs}, but our work differs in two ways. Firstly, while they only applied the principle to the informed RRT*, we apply it to both versions of the informed RRT and the informed RRT*, and analyse their convergence rates. Secondly, we apply multiple path optimisers, presented next, and analyse their effectiveness.

\section{Path Optimisers}
\label{sec:Path_Optimisers}

A path optimiser reduces the cost of a path. In this work, a path's cost is given by its length. A path is represented by a series of $n$ tree nodes, $\mathbf{Q} = \mathbf{q}_0, \mathbf{q}_1, \mathbf{q}_2, \ldots, \mathbf{q}_{n-1}$, such that $\mathbf{q}_0 = \mathbf{q}_I$, $\mathbf{q}_{n-1} = \mathbf{q}_{G}$, and every $(\mathbf{q}_i, \mathbf{q}_{i+1})$ is connected by an edge. The path optimisers we use are described next.

\subsection{Shortcut-based Path Optimisers}
\label{sec:Shortcut_based_Path_Optimisers}

\begin{figure*}[t]
\centering
	\subfloat[path pruning.]{%
		\includegraphics[scale =0.91]{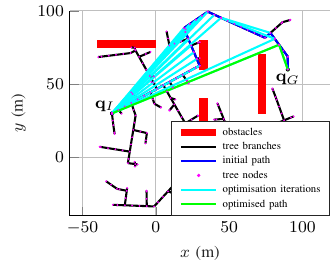}
   	}%
   	\subfloat[random shortcut.]{%
		\includegraphics[scale =0.91]{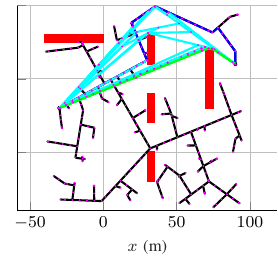}
   	}%
   	\subfloat[The wrapping process.]{%
		\includegraphics[scale =0.91]{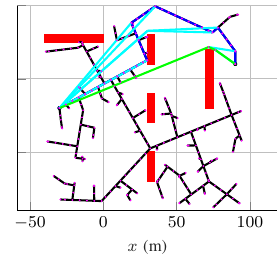}
   	}%
   	\subfloat[GB path optimisation.]{%
   		\includegraphics[scale =0.91]{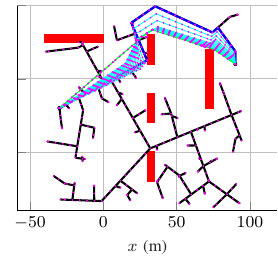}
	}
\captionsetup{justification = centering}
\caption{Example application of the shortcut-based optimisers ((a) to (c)) and the GB path optimiser ((d)).}
\label{fig:exampleApplicationOfShortcutBasedPathOptimisers}
\end{figure*}

Shortcut-based path optimisers iteratively pick two non-consecutive path nodes and attempt replacing the path between them with a shorter direct path. The shortcut-based optimisers we use are briefly described next.

\subsubsection{Path Pruning}
\label{sec:Path_Pruning}

Starting from the first node, $\mathbf{q}_0$, path pruning~\citep{Geraerts_2007} visits each node, $\mathbf{q}_{i}$, and removes the next node, $\mathbf{q}_{i+1}$, if it is possible to connect $\mathbf{q}_i$ directly to $\mathbf{q}_{i+2}$ with a collision-free \emph{shortcut}. Its main drawback is its dependence on the existence of directly-connectible non-consecutive path nodes.

\subsubsection{Random Shortcut}
\label{sec:Random_Shortcut}

Like path pruning, random shortcut (RS)~\citep{Geraerts_2007} works by removing redundant path nodes. Firstly, RS differs in that it introduces additional nodes along the path through discretisation. This removes the dependence on the existence of directly-connectible nodes in the initial path. Secondly, it considers shortcuts randomly, allowing it to cope with a path that has many nodes after discretisation.

\subsubsection{Wrapping Process}
\label{sec:Wrapping_Process}

The wrapping process~\citep{Kim_2015} combines path pruning with a technique that moves non-redundant path nodes towards obstacles so that the resulting path \emph{wraps} around obstacles. Starting from the first node, $\mathbf{q}_0$, it proceeds towards the last node, $\mathbf{q}_{n-1}$, either removing node $\mathbf{q}_{i+1}$, if redundant, or moving it towards obstacles so that the subpath from $\mathbf{q}_i$ to $\mathbf{q}_{i+1}$ wraps onto obstacles.

Fig.~\ref{fig:exampleApplicationOfShortcutBasedPathOptimisers}~((a) to (c)) illustrates the application of the three shortcut-based path optimisers in shortening RRT paths.

\subsection{Gradient-based (GB) Path Optimisation}
\label{sec:Gradient_based_Path_Optimisation}

A gradient-based (GB) path optimiser uses gradient information to reduce path length. We draw on \citet{Campana_2016}'s algorithm, which is summarised next. 

The cost function for path length is computed as:

\vspace*{-1.2em}
\begin{equation}
  L(\mathbf{Q}) = \frac{1}{2}\sum_{k=1}^{n-2} \lambda_{k-1}\Vert \mathbf{q}_k - \mathbf{q}_{k+1} \Vert_W^2,
  \label{eqn:gb_path_optimisation_cost_function_}
\end{equation}

where the coefficients, $\lambda_{k-1}$, are used to cope with obstacles, and are chosen so as to keep the ratio between the lengths of path segments of the optimised path the same as at the initial path. Without the coefficients, the optimised path's nodes are equidistantly spaced -- an unlikely case in the presence of the obstacles. Path length is then minimised using a second-order update rule:

\vspace*{-1.2em}
\begin{equation}
  \mathbf{Q}_{i+1} = \mathbf{Q}_i - \alpha_iH^{-1}\nabla L(\mathbf{Q}_i)^T,
\end{equation}

where $\mathbf{Q}_{i+1}$ denotes new path nodes, $\mathbf{Q}_i$ represents previous path nodes, $\alpha_i$ is the descent step length, and $\nabla L$ and $H$ are the cost function's gradient and Hessian. Without constraints, $\alpha_i = 1$ results in convergence in one step. This theoretical minimum is assumed to contain collisions. So, the algorithm starts with $\alpha_i < 1$ instead. It moves in small steps towards the minimum until a collision is detected, at which point a linearised collision constraint is computed and added to a constraint Jacobian matrix. It then switches to $\alpha_i = 1$, attempting to reach the minimum in one step under the new set of constraints. If this results in collision, it reverts to small steps ($\alpha_i < 1$) until a new collision is detected, resulting in a new collision constraint. The algorithm then switches to $\alpha_i = 1$, attempting to reach the minimum in one step under the new set of constraints. This is repeated until convergence. Figure~\ref{fig:exampleApplicationOfShortcutBasedPathOptimisers}~(d) shows an example application of the algorithm.

\section{Optimised Informed RRTs and Results}
\label{sec:Accelerated_Path_Planners}

We consider a dozen optimised informed RRTs, each resulting from applying one of the path optimisers to one of the benchmark path planners. In the optimised informed path-planning framework, an optimiser cannot be allowed to run indefinitely; it must quickly optimise the path and return control to the planner. This is straightforward for path pruning, the wrapping process, and GB. For RS, however, a \emph{time limit} must be chosen; we use a linear regression model, trained on the convergence times of the other optimisers, for this purpose. Note that all optimisers may be terminated \emph{anytime}, should planning time elapse. 

Fig.~\ref{fig:benchmarkPathPlannersPercentilePlots} shows the convergence patterns for the benchmark path planners and their optimised versions, namely optimised informed basic RRT (OIB-RRT), optimised informed k-nearest RRT (OIKN-RRT) and optimised informed RRT* (OI-RRT*). We first analyse the benchmark path planners, which provide a baseline for the analyses of optimised informed RRTs that follow thereafter. All analyses are based on results from 400 independent runs of each planner, with a maximum of 30 seconds per run, and no upper bound on the number of iterations. In each run, a planner finds an initial path and improves it with the remaining time. The costs of paths found in each run represent the convergence pattern for that run. An infinite cost denotes that no path has been found yet. The algorithms share the same libraries for all common functions, allowing computation time\footnote{All experiments were run in Ubuntu 14.04 on an Intel Celeron(R) B815 CPU with 4GB RAM and 1.60GHz x 2 processor speed.} comparisons.

\subsection{Analyses of Benchmark Path Planners}
\label{sec:Analyses_of_the_Benchmark_Path_Planners}

\begin{figure*}[h]	
   	\includegraphics[scale=1.42]{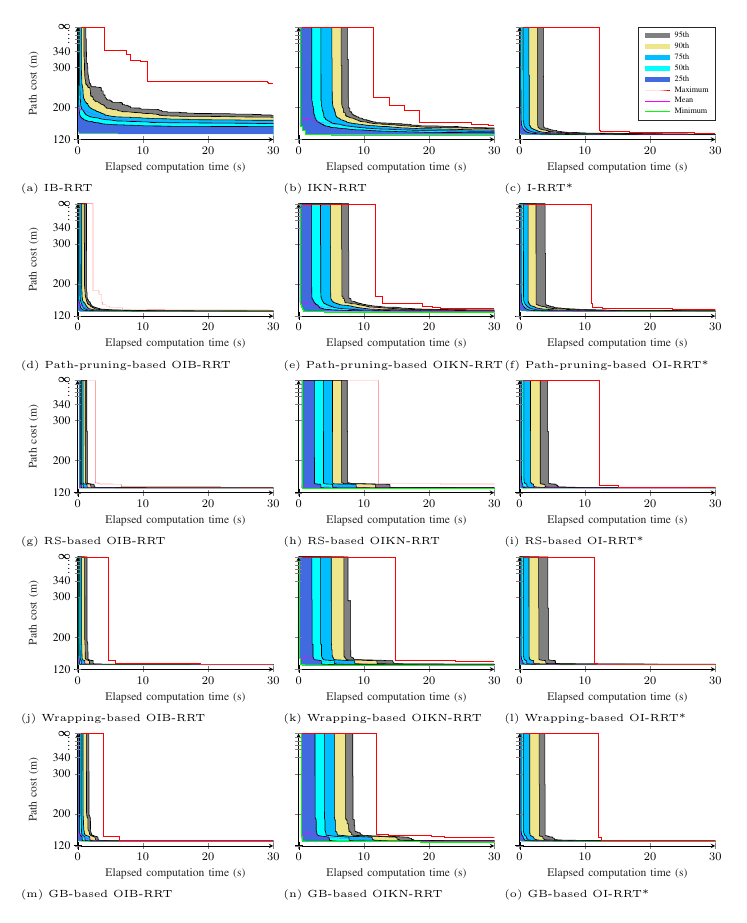}
\caption{Convergence patterns for the benchmark path planners and ((a) to (c)) their optimised versions((d) to (o)).}
\label{fig:benchmarkPathPlannersPercentilePlots}
\end{figure*}

Analysing the convergence patterns\footnote{We analyse two intervals: when planning time is limited ($t \leq 10s$), and when there is more time available for planning ($t = 30s$).} of the benchmark path planners (Fig.~\ref{fig:benchmarkPathPlannersPercentilePlots}~(a) to (c)) helps us confirm that our implementations conform to the expected head-to-head performance. IB-RRT finds paths in all runs when planning time is limited ($t \leq 10s$), affirming that it quickly finds paths, but is unable to improve them even with more planning time ($t = 30s$), as shown by the slow cost decay. The RRT-based benchmark path planners have much higher variability in path costs than I-RRT*; IB-RRT has the most variability. IB-RRT's high suboptimality is confirmed by high costs in all cost categories. Meanwhile, I-RRT*'s maximum and all percentiles converge towards the minimum, while for the RRT-based benchmark path planners this is not the case. These results confirm the asymptotic optimality of I-RRT* and lack thereof in the RRT-based benchmark path planners, as well as the swiftness of RRT-based path planners, especially IB-RRT.

\subsection{Analyses of Optimised Informed RRTs}

The convergence patterns for optimised informed RRTs based on path pruning, RS, the wrapping process and the GB path optimiser (Fig~\ref{fig:benchmarkPathPlannersPercentilePlots}~(d) to (o)) show that, like their benchmarks, all OIB-RRT versions quickly find initial paths in all runs when planning time is limited. Unlike the benchmarks, however, they quickly improve paths given more planning time. When planning time is limited, this new capability gives OIB-RRT a clear edge over the other two optimised path planners which, despite being quick to improve paths, take longer to find initial paths, just like their benchmarks. Analyses are thus limited to the OIB-RRT versions in this case. When more planning time is available, contention is between OIB-RRT and OI-RRT*; analyses are thus limited to these two, in this latter case.

In the first case, among the OIB-RRT versions, the path-pruning-based version is the quickest to find initial paths in all runs (within 2.36 $s$), followed by the RS-based version, which takes slightly more time (2.70 $s$). At finding paths in all runs, RS-based OIB-RRT's worst cost is 21.5\% smaller than path-pruning-based OIB-RRT's. RS-based OIB-RRT's path costs are also less variable than path-pruning-based OIB-RRT's. Since RS-based OIB-RRT only lags slightly, attains a much less worst-case cost on catching up, and has less variability, it can be considered to edge the latter when planning time is limited. The wrapping process and the GB path optimiser add significant computation time for OIB-RRT. For the wrapping process, since IB-RRT paths typically have more segments, wrapping segments onto obstacles takes more time. Similarly, the GB path optimiser spends more time on collision checking. 

In the second case, i.e. when there is more planning time, the costs and variability of all the OIB-RRT and  OI-RRT* versions at the end of the computation period ($t = 30s$) are used as the basis for comparison. Among the OIB-RRT versions, the RS-based OIB-RRT has the least costs in all cost categories, and has the least variability. It is followed by the wrapping-based OIB-RRT, then the GB-based OIB-RRT, and last comes the path-pruning-based OIB-RRT. Then among the OI-RRT* versions, the GB-based OI-RRT* comes first, and is closely followed by the RS-based one. Third comes the wrapping-based OI-RRT*, and the path-pruning-based OI-RRT* is last. Interestingly, when the OIB-RRT versions are compared with the OI-RRT* versions, the RS-based and the wrapping-based OIB-RRT outperform all the OI-RRT* versions when there is more planning time; the GB-based and the path-pruning-based OIB-RRT respectively have comparable performance to the GB-based OI-RRT* and the RS-based OI-RRT* -- the best two OI-RRT* versions.

\section{Conclusion}
\label{sec:Conclusion} 

This paper investigates the use of a combination of informed sampling and path optimisation to accelerate convergence of RRT* and RRT-based path planners. The key goal is to ascertain if incorporating informed sampling and path optimisation can help a quick, but almost-surely suboptimal, RRT-based planner attain comparable or better performance than an asymptotically optimal RRT*-based counterpart. Two RRT-based path planners, namely informed basic RRT and informed k-nearest RRT, that make use of informed sampling, are used as benchmarks, along with informed RRT*. Four path optimisers, namely path pruning, random shortcut, the wrapping process and a gradient-based path optimiser, are used to accelerate convergence of the benchmark path planners, resulting in a family of algorithms known as \emph{optimised informed RRTs}. Analyses show that when planning time is limited, all optimised informed basic RRT (OIB-RRT) versions outperform all optimised informed RRT* (OI-RRT*) versions. The best two OIB-RRT versions (RS-based and wrapping-based) also outperform all OI-RRT* versions when there is more planning time; the last two OIB-RRT versions (GB-based and path-pruning-based) have comparable performance to the best two OI-RRT* versions in this case. Thus incorporating informed sampling and path optimisation does help the quick, but almost-surely suboptimal, basic RRT attain better performance than its RRT*-based counterpart.

\end{document}